# An Empirical Analysis of Likelihood-Weighting Simulation on a Large, Multiply Connected Belief Network

## Michael Shwe and Gregory Cooper

### Section on Medical Informatics, Medical School Office Building X215, Stanford University, Stanford, CA 94305-5479

## Abstract

We analyzed the convergence properties of likelihood-weighting algorithms on a two-level, multiply connected, belief-network representation of the QMR knowledge base of internal medicine. Specifically, on two difficult diagnostic cases, we examined the effects of Markov blanket scoring, importance sampling, and self-importance sampling, demonstrating that the Markov blanket scoring and self-importance sampling significantly improve the convergence of the simulation on our model.

## 1. Introduction

The Quick Medical Reference (QMR) program is a decision-support tool for diagnosis in internal medicine that was developed at the University of Pittsburgh as the successor to INTERNIST-1 [Miller, Pople, et al., 1982]. Designed to assist a physician in making difficult diagnoses, QMR is built on one of the largest knowledge bases (KBs) in existence. We are developing the foundation for a decision-theoretic version of QMR, which we call QMR-DT for *Quick Medical Reference—Decision Theoretic*.

Our research to date has focused on building the QMR-DT KB, a probabilistic reformulation of the QMR KB, and on developing a method for inference on the QMR-DT KB. In this paper, we concentrate our discussion on likelihood weighting as a method of inference. Before describing the likelihood-weighting algorithm, we briefly examine the QMR-DT KB.

## 2. The QMR-DT Model

We have reformulated the associations between diseases and findings of the INTERNIST-1 disease profiles [Miller, Pople, et al., 1982] into a belief-network representation.[1] This reformulation is described in [Heckerman, 1989; Henrion, 1988; Shwe, Middleton, et al., 1990]. The QMR-DT KB consists of a two-level belief network of $n$ diseases and $m$ findings, as shown in Figure 1.

---

[1] We are currently using the INTERNIST-1 KB (circa 1986), rather than the more recent QMR KB. These two KBs are quite similar, to the extent that the methods in this paper are applicable to transforming the latter KB as well. For simplicity, where the distinction between the INTERNIST-1 KB and QMR KB is inconsequential, we will refer to the INTERNIST-1 KB as the QMR KB

Each of the $n$ diseases $\{D_1,...,D_n\}$ may be present or absent in a patient, and each of the $m$ findings $\{F_1,...,F_m\}$ may be unobserved or observed to be present or absent. We define $d_i$ as the state of disease $D_i$. If $D_i$ is present, then $d_i$ = present; if $D_i$ is absent, then $d_i$ = absent. We use $P(D_i | \varepsilon)$ as a shorthand for $P(d_i$ = present $| \varepsilon)$, where $\varepsilon$ is any evidence observed. We refer to a disease hypothesis $H$ as an assignment of present or absent to each disease in $\{D_1,...,D_n\}$, that is,

$$H = \bigcup_{i=1}^{n} d_i$$

Also, we define $H^+$ to be the set of diseases $D_i$ such that $d_i$ = present in $H$.

Let $N_F$ be a set of findings observed for a particular patient, where $N_{F^+}$ is the set of findings observed to be present and $N_{F^-}$ is the set of findings observed to be absent. Note that many findings may be unobserved and thus appear in neither $N_{F^+}$ nor $N_{F^-}$. We define $f_j$ as the state of finding $F_j$.

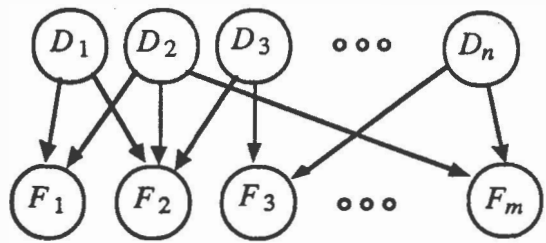

Figure 1 The two-level belief-network representation of the current QMR-DT KB. The disease nodes are labeled $D_1,...,D_n$ and the finding nodes are labeled $F_1,...,F_m$. The probabilistic dependencies between diseases and findings are specified with directed arcs between nodes, where an arc points in the causal direction that we assume; that is, we assume that diseases cause findings.

An arc of probabilistic dependency between nodes representing a disease $D_i$ and finding $F_j$ exists in the QMR-DT KB if and only if there exists a link between $D_i$ and $F_j$ in the QMR disease profile of $D_i$. Disease-to-disease dependencies are not modeled presently in the QMR-DT KB. The current QMR-DT KB contains $n$ =

534 adult diseases and $m = 4040$ findings, with 40,740 arcs depicting disease-to-finding dependencies.

To reduce the representational and computational complexity of QMR-DT, we made several simplifying assumptions. Assumptions evident from Figure 1 include marginal independence of diseases, conditional independence of findings given any hypothesis of diseases, and the assumption that findings are manifestations of disease. Also, we assume that diseases and findings are binary valued. The assumption that diseases are marginally independent allows us to calculate $P(H)$ by

$$P(H) = \prod_{i=1}^{n} P(d_i) \quad . \tag{1}$$

The assumption that findings are conditionally independent given any disease hypothesis allows us to calculate $P(N_F | H)$ as

$$P(N_F | H) = \prod_{F_j \in N_F} P(f_j | H) \quad . \tag{2}$$

We model the influence of multiple diseases on a finding assuming causal independence in a noisy-OR gate interaction [Pearl, 1988]. Under the assumption of a noisy-OR gate, we can avoid representation of the full set of conditional probabilities of the state of a finding given each possible state of the finding's parents. Consider a belief network with binary finding $F_j$ where $F_j$ has binary parents $D_1, D_2, ..., D_k$. To construct the complete conditional probability table associated with the arcs from $D_1, D_2, ..., D_k$ to $F_j$, we would need to acquire a conditional probability for each of the $2^k$ states of the parents of $F_j$. If we assume causal independence, we need to acquire only $k$ conditional probabilities of the form $P(F_j | \text{only } D_i)$,[2] where $1 \le i \le k$. These conditional probabilities are derived from a mapping of QMR frequencies to probabilities [Shwe, Middleton, et al., 1990]. This mapping was assessed from Randy Miller, one of the primary developers of INTERNIST-1 and QMR. The results of the mapping appear in Table 1.

Assuming a noisy-OR gate interaction among diseases on a finding, we compute $P(f_j | H)$ as

$$P(f_j | H) = 1 - \prod_{\pi(F_j) \cap \{D_i : D_i \in H^+\}} [1 - P(f_j | \text{only } D_i)] \quad , \tag{3}$$

where $\pi(F_j)$ are the diseases that are parents of $F_j$.

In addition to obtaining the conditional probabilities relating findings to diseases, we derived prior probabilities on diseases in the QMR-DT KB from data compiled by the National Center for Health Statistics (NCHS) on approximately 192,000 inpatients discharged from short-stay nonfederal hospitals in 1984 [Lawrence, 1986].

Table 1 A mapping between QMR frequencies and probabilities

| QMR frequency | $P(F_j | \text{only } D_i)$ |
|---|---|
| 1 | 0.025 |
| 2 | 0.20 |
| 3 | 0.50 |
| 4 | 0.80 |
| 5 | 0.985 |

## 3. Algorithms for Inference

Given a set of positive and negative findings $N_F$ and a model of the dependencies between diseases and findings in internal medicine, our goal is to compute $P(D_i | N_F)$, the posterior marginal probability for each disease $D_i : 1 \le i \le n$. We contrast $P(D_i | N_F)$ with $P(\text{only } D_i | N_F, \mu)$, where $\mu$ is the assumption that diseases are mutually exclusive. This assumption is clearly not applicable to the general problem of diagnosis in internal medicine, where patients often have several diseases simultaneously. The posterior marginal probability, on the other hand, implicitly acknowledges the possibility of there being one or more diseases in a patient. In the next subsection, we first discuss the complexity of calculating $P(\text{only } D_i | N_F, \mu)$, and then describe the complexity of calculating $P(D_i | N_F)$.

### 3.1 Exact Algorithms

We refer to Bayes' rule under the assumptions of single-disease hypotheses and conditional independence of findings as *tabular Bayes' rule*[3]:

$$P(\text{only } D_i | N_F, \mu) = \frac{P(N_F | \text{only } D_i) P(\text{only } D_i)}{\sum_{k=1}^{n} P(N_F | \text{only } D_k) P(\text{only } D_k)} \tag{4}$$

---

[2] We distinguish $P(F_j | \text{only } D_i)$ from $P(F_j | D_i)$, where the former denotes the probability of the event that $F_j$ occurs given that only $D_i$ occurs, and that, for all $k \ne i$, $D_k$ is absent. By contrast, we use the notation $P(F_j | D_i)$ to mean the probability of the event that $F$ occurs given that $D_i$ occurs and that for all $k \ne i$, each $D_k$ occurs based on its prior probability.

[3] The name *tabular Bayes'* rule is derived from the notion that we can compute $P(\text{only } D_i | N_F, \mu)$ as in Equation 4 from a $n \times m$ table of probabilities of the form $P(f_j | \text{only } D_i)$, where $1 \le i \le n$ and $1 \le j \le m$.

Although the single-disease assumption is extremely restrictive, the tabular Bayes' formulation is appealing because of its low degree of computational complexity, $O(nm)$.

Consider generalizing to allow the diagnostic hypothesis $H$ to contain any subset of diseases in the KB. This generalization is consistent with the QMR-DT model. Straightforward application of Bayes' rule to the QMR-DT two-level belief network yields an inferential complexity of $O(m'n2^n)$ [Shwe, Middleton, et al., 1990]. The $2^n$ term arises since the denominator must be summed over $2^n$ disease hypotheses:

$$P(d_i \mid N_F) = \frac{\sum_{H:D_i \in H^+} P(N_F \mid H) P(H)}{\sum_H P(N_F \mid H) P(H)} . \quad (5)$$

Suppose that we were to perform inference using Equation 5 on machinery that could support 100 billion multiplications per second. By comparison, a Cray Y-MP/832 with eight processors has a limit of 2.7 billion floating-point operations per second. Consider the time it would take to compute only the terms $P(H)$ in the summation of the denominator of Equation 5. Each $P(H)$ term requires 534 multiplications. Thus, we would need more than $2^{534} \times 534 \approx 3 \times 10^{163}$ multiplications to compute the denominator of Equation 5. On our hypothetical machinery, it would take more than $10^{144}$ years to complete the computation. Clearly, the brute-force application of Bayes' theorem to inference on the QMR-DT belief network is impractical. Moreover, the problem of probabilistic inference on two-level binary-valued belief networks such as QMR-DT is known to be NP-hard [Cooper, 1990]. Accordingly, we have sought to develop special-case algorithms [Heckerman, 1989] and approximation algorithms [Henrion, 1988] to perform more efficient inference on the QMR-DT belief network.

### 3.2 Approximation Algorithms

Because the general problem of probabilistic inference on belief networks is NP-hard, we have focused on developing *approximation algorithms*. The approximation algorithms we have explored compute estimates of the posterior marginal probabilities of diseases that converge in the limit to the true posterior marginal probabilities, given the QMR-DT model.

We have implemented an approximation algorithm called *likelihood weighting*, which places no a priori restrictions on the connectivity of the belief network. Our goal is to investigate the performance of likelihood weighting on the current QMR-DT belief network and then to use the algorithm on future versions of the network that contain a richer collection of dependencies (for example, dependencies among diseases). Likelihood weighting, a stochastic simulation algorithm, has been described by Fung and Chang [Fung & Chang, 1989] and by Shachter and Peot [Shachter & Peot, 1989].

The likelihood-weighting scheme instantiates each nonevidence node based on the state of the node's parents and computes efficiently the likelihood of the instantiated state of the unobserved nodes given the evidence. The likelihood score, called the *sample score*, is then added to a slot for each event that occurs in the trial. After a specified number of trials, we estimate the probability of an event by dividing the aggregate sample scores of the event by the total aggregate sample scores.

More formally, in the case of the two-level QMR-DT belief network described in Section 2, we estimate the marginal posterior probability of Equation 5 by

$$\widehat{P}(d_i \mid N_F) = \frac{\sum_{j=1}^{t} Z(H_j \mid N_F) U(d_i, H_j)}{\sum_{j=1}^{t} Z(H_j \mid N_F)} , \quad (6)$$

where $H_j$ is the state of all the diseases as instantiated in the $j$th trial, $t$ is the total number of trials, $Z(H_j \mid N_F)$ is the sample score for the $j$th trial, and $U(d_i, H)$ is 1 if the value of $D_i$ in $H$ is $d_i$, and is 0 otherwise. The sample score is given by

$$Z(H_j \mid N_F) = \frac{P(N_F \mid H_j) P(H_j)}{P'(H_j)} \quad (7)$$

where $P'$ is the sampling distribution. In the simplest case, we instantiate the disease nodes based on their prior probabilities; that is, $P' = P$.

Alternatively, we can focus the likelihood-weighting simulation on certain instantiations of the network. Using this technique, called *importance sampling*, the algorithm instantiates the diseases not by their prior probabilities $P(D_i)$, but rather by any sampling distribution $P'(D_i)$ [Shachter & Peot, 1989]. The only restriction on $P'$ is that $P'(d_i) > 0$ whenever $P(d_i) > 0$. The simulation's estimates of the posterior distribution will converge in the limit of infinity to the true posterior distribution as long as $P'$ follows this restriction [Rubinstein, 1981]. The estimates will converge most quickly when $P'(D_i)$ is equal to the true posterior distribution $P(D_i \mid N_F)$ for each $D_i$, where $1 \leq i \leq n$. Of course, if we knew the true posterior distribution, then we would not have to perform the simulation. We can attempt to approximate the true posterior distribution using any method of our choosing—a heuristic method, for example—to improve the convergence of the simulation. We can also update $P'$ based on the simulation's current estimates of the posterior distribution. This technique is called *self-importance sampling* [Shachter & Peot, 1989].

501

The self-importance updating function that we use is

$$P'_{new}(d_i) = \frac{P'_0(d_i) + g(t) \widehat{P}_{current}(d_i | F)}{g(t) + 1}, \quad (8)$$

where $g(t)$ is a linear function of the number of trials and $P'_0$ is the original sampling distribution. We use $P'_0$ in the updating function so that very early in simulation the update will not converge to extreme probabilities (that is, close to 0 or 1). We use the set of likely diseases generated by a heuristic algorithm to set $P'_0$. This heuristic algorithm, which we call the *iterative tabular Bayes'* algorithm (ITB), applies successively the tabular Bayes' calculation to various subsets of the observed findings $N_F$; see [Shwe, Middleton, et al., 1990] for a detailed description of this algorithm. We refer to the set of diseases generated by the ITB algorithm as the *heuristic-importance set*, which contains approximately 25 diseases.

We set the original importance distribution such that the expected number of diseases instantiated from the heuristic-importance set is 1. That is, for all $D$ in the heuristic-importance set, $P'_0(D_i) = 1/N$, where $N$ is the cardinality of the heuristic importance set. We then set $P'_0(D_i)$ for all $D_i$ not in the heuristic-importance set to the greater of $10^{-3}$ or the prior probability on $D_i$. We use the threshold of $10^{-3}$ so that we can expect each disease to be instantiated during simulation a number of times before the sampling distribution is updated based on the simulation's probability estimates. For example, suppose that we update the sampling distribution after the first 10,000 trials of simulation. Then, we would expect each disease not in the heuristic-importance set to be instantiated $10,000 \times 10^{-3} = 10$ times.

In addition to using importance sampling and self-importance sampling to improve the convergence properties of the simulation, we use Markov blanket scoring [Shachter & Peot, 1989]. To apply the Markov blanket scoring modification to the QMR-DT belief network, we add a fraction of the sample score to slots for both the present and the absent states of each disease. For a specific disease, this fraction of the sample score is proportional to the probability of each state of the disease given the state of the rest of the network, or—as Pearl demonstrates in [Pearl, 1987]—the state of the disease node's Markov blanket. To incorporate Markov blanket scoring into Equation 6, we simply redefine the function $U$ such that

$$U(d_i, H_i) = P(d_i | w_{D_i}), \quad (9)$$

where $w_{D_i}$ is the state of all the variables in the network except $D_i$. We can compute $P(d_i | w_{D_i})$ efficiently using the Markov blanket of $d_i$:

$$P(d_i | w_{D_i}) \propto P(d_i) \prod_{F_j \in \{S(D_i) \cap N_F\}} P(f_j | w_{D_i}, d_i), \quad (10)$$

where $S(D_i)$ are the children of $D_i$.

Note that the complexity of calculating the sample score of Equation 7 is $O(|N_F|n)$. If, during the computation of the sample score, we store the values of $P(f_j | H)$, we can compute $P(d_i | w_{D_i})$ in $O(|S(D_i)|)$ additional time. In the worst case, where each $D_i$ is connected to each $F_j \in N_F$, we require $O(|N_F|n)$ additional time to compute the Markov blanket probabilities for all the diseases in the network.

In summary, our simulation algorithm incorporates an importance distribution from a heuristic algorithm, self-importance sampling, and Markov Blanket scoring. We shall refer to this algorithm as $S$. In Section 4, we discuss among other experiments, a study of the convergence properties of S.

## 4. Experimental Design

We examined the convergence properties of the S algorithm on two test cases. Also, in a sensitivity analysis of components of S, we examined the effect on convergence of importance-sampling from a heuristically derived set of diseases, self-importance sampling, and Markov blanket scoring. We implemented and evaluated S and the variations of S in LightSpeed Pascal, version 2.03, on a Macintosh IIci.

### 4.1 Test Cases

We used two diagnostic cases abstracted from published clinicopathological conference (CPC) exercises. CPCs represent some of the more difficult diagnostic challenges to physicians, often containing many positive manifestations and multiple diseases in the diagnosis, which is determined by pathological investigation at autopsy. The two cases we used in this study appeared in [Cryer & Kissane, 1974] and [Castleman, Scully, et al., 1972]. We shall refer to these cases as CPC1 and CPC2, respectively. Both of these cases were abstracted by the INTERNIST-1 group for testing of the INTERNIST-1 system. Information on the test cases appears in Table 2.

We used CPC1 while developing the S algorithm, but did not use CPC2 prior to the study reported here. We selected CPC1 from the set of CPCs that we had run previously because of the multiple diseases in the diagnosis and the large number of positive manifestations. We selected CPC2 randomly from a set of four test cases, each of which met our criteria of greater than or equal to 30 positive findings, 10 negative findings, and three diseases in the diagnosis. We used these criteria to select one of the more difficult CPCs.

Table 2  Test cases

|  | Number of diseases | | |
|---|---|---|---|
| Case | in diagnosis | $|N_{F^+}|$ | $|N_{F^-}|$ |
| CPC1 | 5 | 51 | 2 |
| CPC2 | 3 | 50 | 24 |

### 4.2 Reference Distributions

Because we do not know of any practical algorithm that can compute the exact posterior marginal probabilities of disease using the QMR-DT belief network, we use S to create a reference distribution for CPC1 and CPC2. We henceforth refer to the 64-hour reference runs of S on a Macintosh IIci as the REF algorithm. The output from REF represents our best estimates of the true posterior marginal probabilities of disease, given the QMR-DT belief network.

Although we are not certain that the estimates provided by the REF algorithm are reasonably close to the posterior probabilities implied by the QMR-DT model, we observed during our development of S that successive, independent 64-hour runs of S produced similar probability distributions. In Section 4.3, we describe the metric we used for comparing probability distributions.

### 4.3 Convergence Metric

To compare probability distributions over the set of diseases in the QMR-DT KB, we used a correlation coefficient. Let $A(i)$ be the disease number for the disease that is ranked as the $i$th most probable disease by algorithm $A$. Thus, for example, $D_{A(1)}$ is the most probable disease according to algorithm $A$. Let $P_X(D_{A(i)} | N_F)$ be the probability that algorithm $X$ assigns to disease $D_{A(i)}$ given the finding set $N_F$. We define the correlation $r(A, B)$ as the correlation coefficient over the pairs $(P_A(D_{A(i)} | N_F), P_B(D_{A(i)} | N_F))$ for $1 \leq i \leq 20$. In general, $r(A, B)$ is not symmetric. In the analysis of the convergence properties of S and variations of S, we assign $A$ to be REF and $B$ to be S or one of the modified versions of S that we describe in Section 4.4.

### 4.4 Sensitivity Analysis

We examined the sensitivity of the simulation algorithm to its component heuristics by comparing the probabilistic output of three modified versions of S to the probabilistic output of REF. We refer to the S algorithm with no self-importance sampling as *S/NSI*. Similarly we refer to the S algorithm with no ITB initialization heuristic as *S/NITB*. Like S, S/NSI obtains its initial importance distribution $P'_0$ from the heuristic-importance set of ITB. (For all $D_i$ in the heuristic-importance set, $P'_0(D_i) = 1/N$, where $N$ is the cardinality of the heuristic- importance set.) By contrast, S/NITB does not use the heuristic-importance set to generate $P'_0$. Rather, S/NITB sets $P'_0(D_i)$ for all $D_i$ to the greater of $10^{-3}$ or the prior probability on $D_i$. Finally, we refer to the S algorithm with no Markov blanket scoring as *S/NMBS*.

### 5. Results and Analysis

We ran REF, S, S/NMBS, S/NITB, and S/NSI on test cases CPC1 and CPC2. Recall that we ran the REF algorithms for 64 hours. We began self-importance sampling after $t = 20,000$ trials, updating $P'$ every 10,000 trials thereafter. Also, we set $g(t) = t/20,000$ (Equation 8) in the REF algorithms. We ran the S, S/NMBS, S/NITB, and S/NSI algorithms for 16 hours on each of the two CPCs. For the S, S/NMBS, and S/NITB algorithms, we began self-importance sampling after $t = 10,000$ trials, updated $P'$ every 5,000 trials thereafter, and set $g(t) = t/10,000$. The ranks and posterior marginal probabilities that REF assigned to the diagnoses of CPC1 and CPC2 appear in Table 3.[4]

The total number of trials completed in each of the runs is listed in Table 4. Note that the Markov blanket scoring modification to the simulation algorithm approximately doubles the time per trial, as our analysis in Section 3.2 suggests.

Table 3  Ranks and posterior marginal probabilities that REF assigned to the diagnoses of two CPC cases

| Disease | Rank | $P(D_i | N_F)$ |
|---|---|---|
| CPC1 | | |
| acute thrombophlebitis of the lower extremities | 2 | 0.99 |
| mitral stenosis | 4 | 0.80 |
| pulmonary infarction | 19 | 0.14 |
| hepatic congestion with centrolobular necrosis | 61 | 0.01 |
| secondary pulmonary hypertension[a] | 211 | 0.00066 |
| CPC2 | | |
| hemiplegia | 1 | 0.99 |
| atheromatous embolism | 2 | 0.99 |
| cerebral embolism | 3 | 0.95 |

[a]REF assigned a rank of 1 and a $P(D_i | N_f) = 0.99$ to *primary* pulmonary hypertension.

---

[4] The REF algorithm estimated an extremely low posterior marginal probability for secondary pulmonary hypertension, whereas the estimate for $P(\text{PRIMARY PULMONARY HYPERTENSION} | N_F)$ was 0.99. We believe that this discrepancy was due largely to the absence of disease-to-disease dependencies in the current QMR-DT model—particularly the absence of an arc between MITRAL STENOSIS and SECONDARY PULMONARY HYPERTENSION. We note that such a link exists in the QMR knowledge base.

503

Table 4  Number of trials completed by simulation algorithms

| Algorithm | Time (minutes) | Number of trials | Trials/minute |
|---|---|---|---|
| CPC1 | | | |
| REF | 3,840 | 450,731 | 117 |
| S | 960 | 103,327 | 108 |
| S/NMBS | 960 | 219,010 | 228 |
| S/NITB | 960 | 113,501 | 118 |
| S/NSI | 960 | 108,650 | 113 |
| CPC2 | | | |
| REF | 3,840 | 283,872 | 74 |
| S | 960 | 71,141 | 74 |
| S/NMBS | 960 | 173,713 | 181 |
| S/NITB | 960 | 75,352 | 78 |
| S/NSI | 960 | 80,059 | 83 |

Table 5  Correlation coefficients $r(REF,B)$ after 960 minutes of simulation

| | Test Case | |
|---|---|---|
| Algorithm | CPC1 | CPC2 |
| S | 0.95 | 0.97 |
| S/NMBS | 0.83 | 0.82 |
| S/NITB | 0.91 | 0.96 |
| S/NSI | 0.14 | 0.54 |

Table 6 Statistics on output from the ITB algorithm

| Test case | \| HIS \|[a] | \| HIS $\cap$ $\{D_{REF(1)},...,D_{REF(20)}\}$ \| |
|---|---|---|
| CPC1 | 30 | 12 |
| CPC2 | 20 | 8 |

[a]HIS: heuristic-importance set from ITB.

The correlation coefficients $r(REF, B)$ where algorithm $B$ is S, S/NMBS, S/NSI, or S/NITB, appear in Table 5. We see that, in both CPC1 and CPC2, the probabilities of the S algorithm correlated most closely with those of the reference distribution. The scatterplot of the 20 pairs of probabilities used to calculate $r(REF,S)$ for CPC1 appears in Figure 2. Figure 3 shows a similar scatterplot for CPC2.

In addition to recording the correlation coefficients after the 960 minutes of simulation time, we recorded $r(REF, B)$ every 5000 trials. Graphs of these correlation coefficients as a function of time appear in Figures 4 and 5. We see that, in both CPC1 and CPC2, the S algorithm is not particularly sensitive to the ITB heuristic, since the probabilities of S/NITB converged to those of the reference distribution nearly as well as did those of S (which did use the ITB heuristic as a starting point). Moreover, the fact that probabilities of S/NITB correlated well with those of the reference distribution increases our belief that the reference distribution is similar to the true posterior probabilities given the QMR-DT model. Recall that S/NITB did not use the heuristic-importance set to set the original sampling distribution $P'_0$, but rather used the prior probabilities of diseases to set $P'_0$. Despite the difference in the initial sampling distributions of S/NITB and REF, the estimates of the posterior distribution from S/NITB converged to those of REF, demonstrating that the posterior estimates of REF are not overly sensitive to the initial sampling distribution.

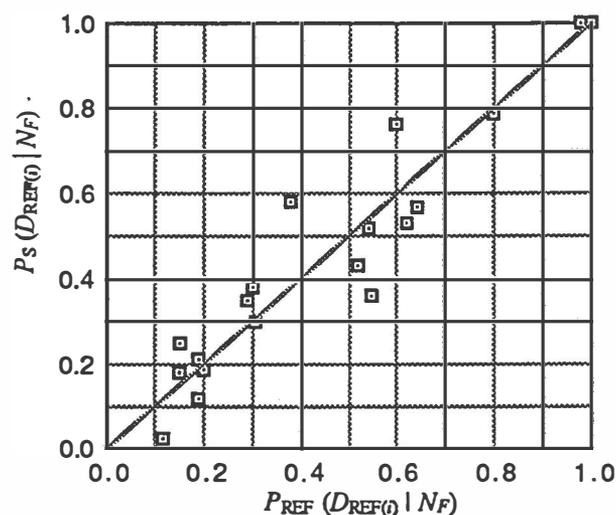

Figure 2  A scatterplot of the probabilities corresponding to $r(REF,S)$ for CPC1.

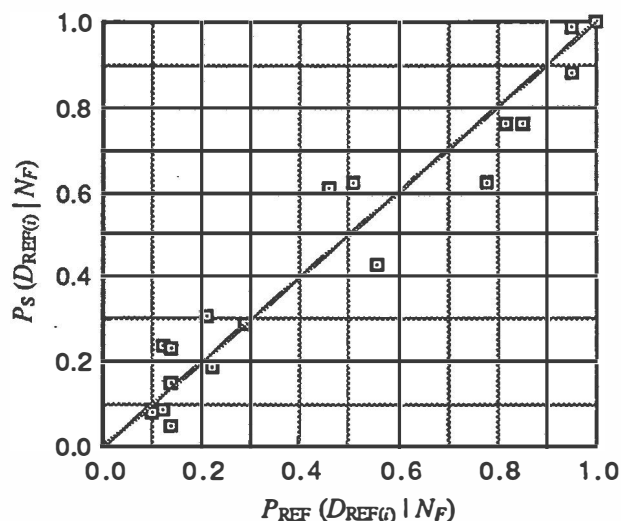

Figure 3 A scatterplot of the probabilities corresponding to $r(REF,S)$ for CPC2.

We might expect the posterior distribution of S/NITB to be slower than that of S to converge to the reference distribution, because S/NITB initially did not bias its sampling distribution toward likely disease candidates. From the convergence on CPC1 depicted in Figure 4, we see that the convergence as a function of time of S/NITB is generally slower than that of S. By contrast, the convergence on CPC2 depicted in Figure 5 reveals that, although the distribution of S was initially better correlated with REF distribution than was the distribution of S/NITB, after about 250 minutes of simulation time, the S/NITB distribution actually converges to REF more quickly than does that of S. We would expect S/NITB to converge more quickly than S if the ITB heuristic were poor; however, ITB performed well. ITB suggested 30 disease candidates for CPC1, 12 of which were ranked in the top 20 diseases of the REF distribution for CPC1. (See Table 6.) On CPC2, ITB suggested 20 disease candidates, eight of which were ranked in the top 20 diseases of REF.

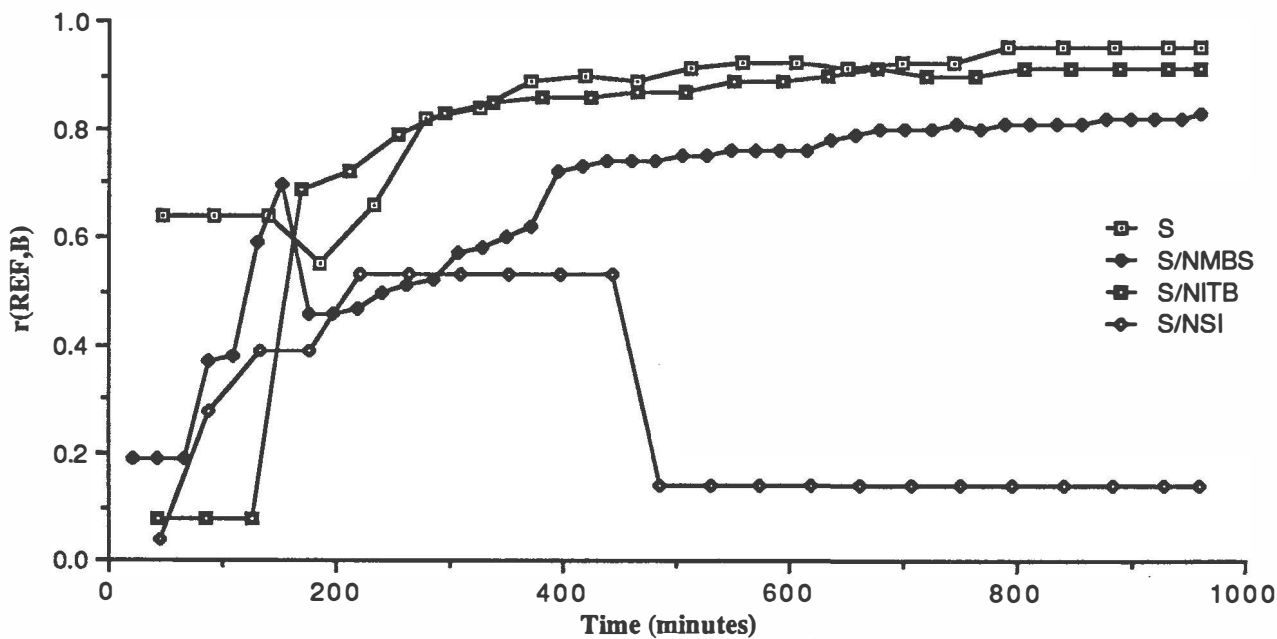

Figure 4  Correlation coefficients $r(REF, B)$ of simulation algorithms as a function of time for CPC1. A correlation coefficient of 1 denotes that the probabilities assigned by REF to the top 20 diseases in the differential were perfectly correlated with the probabilities assigned by algorithm $B$ to those diseases.

Also from Figures 4 and 5, we see that, without Markov blanket scoring, the convergence of the simulation algorithm degrades noticeably. Although the Markov blanket modification more than doubles the computational time per trial, the improvement in convergence to the simulation outweighs the added computational cost. In both CPC1 and CPC2, after 960 minutes, the distribution of S/NMBS did not converge to the reference distribution as well as the distribution of S did. For the REF run on CPC1 and the REF run on CPC2, we recorded the distribution of the sample scores and joint probabilities $P(N_F, H_i)$ generated during the course of the simulation. In Figure 6 appears a distribution of the log of $P(N_F, H_i)$ for the hypotheses $H_i$ generated by REF, where $N_F$ are from CPC1. Note that the hypotheses in the distribution are not unique. The distribution in Figure 6 is similar to that in Figure 7, which depicts the instantiations of S on CPC1. In both Figures 6 and 7, the peak at about $P(N_F, H_i) = 10^{-130}$ represents the null hypotheses (that is, the absence of disease) that were instantiated by the simulation. An important feature of the distributions in Figures 6 and 7 is the absence of outliers on the right-hand side of the curve. (The largest values $P(N_F, H_i)$ for hypotheses instantiated by REF were approximately $10^{-44}$.) An enlargement of the upper tail of the distribution pictured in Figure 7 appears in Figure 8. The presence of outliers typically indicates that the simulation has not yet sampled a sufficient number of hypotheses. Since outliers with high joint probabilities greatly increase the variance in the distribution of the joint probabilities of the instantiations of the network, the outliers also increase the variance in successive estimates of the posterior marginal probabilities of disease.

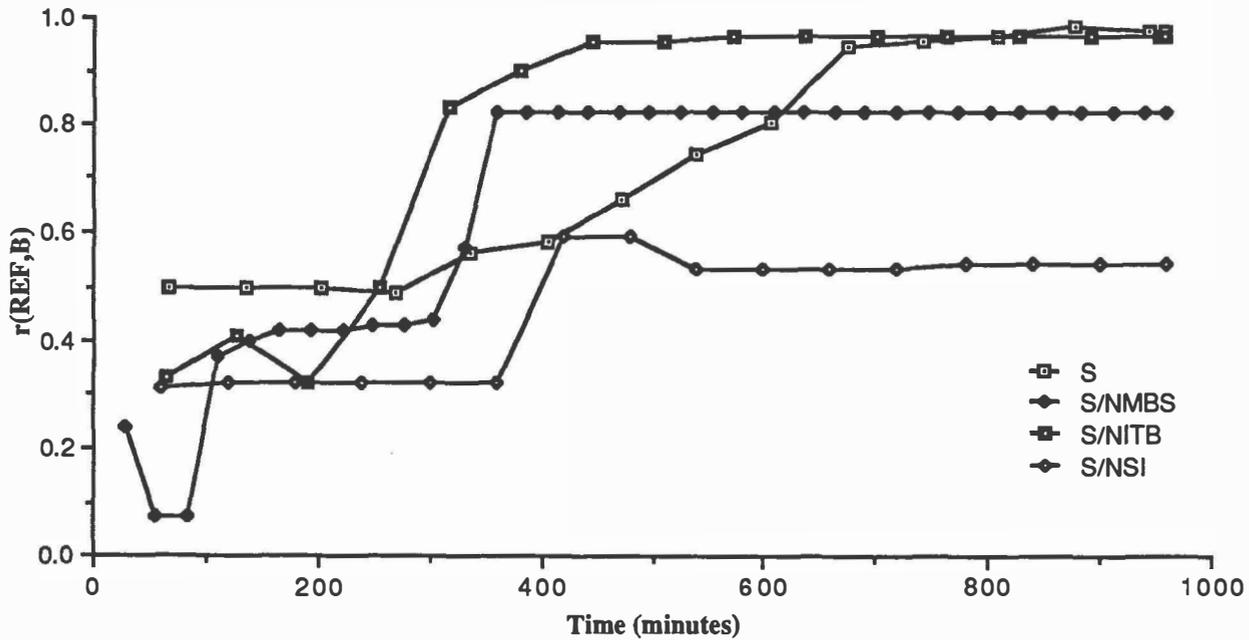

Figure 5 Correlation coefficients $r(REF, B)$ of simulation algorithms as a function of time for CPC2.

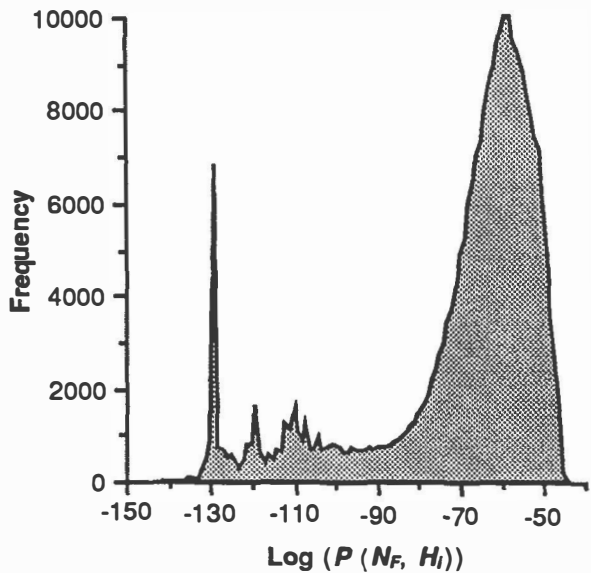

Figure 6 A distribution of the log of the joint probabilities $P(N_F, H_i)$ of the hypotheses $H_i$ instantiated during the REF simulation on the findings $N_F$ of CPC1.

For example, the distribution of $P(N_F, H_i)$ generated by S/NSI on CPC1 contains a number of outliers with large $P(N_F, H_i)$ values. This distribution appears in Figure 9, and an enlargement of the upper-tail end of the distribution appears in Figure 10. Note the two outlier instantiations with $P(N_F, H_i)$ values of about $10^{-57}$. By contrast, the distribution of S on CPC1 (Figures 7 and 8) did not contain such outliers, and the largest values $P(N_F, H_i)$ of hypotheses instantiated were about $10^{-44}$. The largest values $P(N_F, H_i)$ for hypotheses instantiated by REF were about $10^{-44}$. Thus, the S/NSI simulation failed to instantiate many of the disease hypotheses with the largest $P(N_F, H_i)$ values, indicating that either not enough trials were performed or the importance distribution was poor.

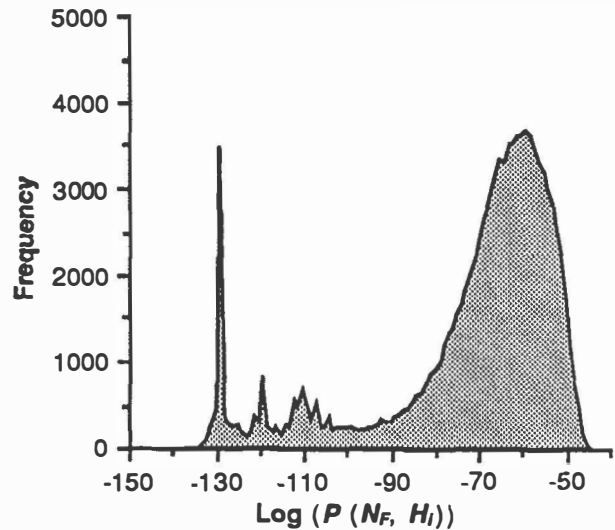

Figure 7 A distribution of the log of the joint probabilities $P(N_F, H_i)$ of the hypotheses $H_i$ instantiated during the S simulation on CPC1.

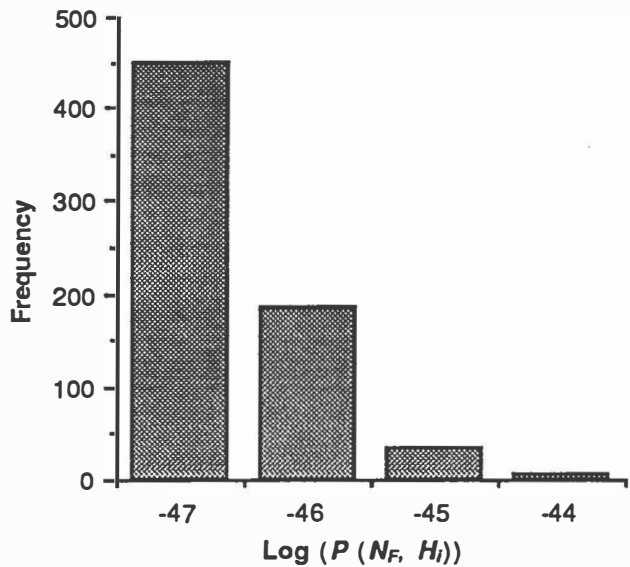

Figure 8 An enlargement of the right-hand side of the distribution that appears in Figure 7.

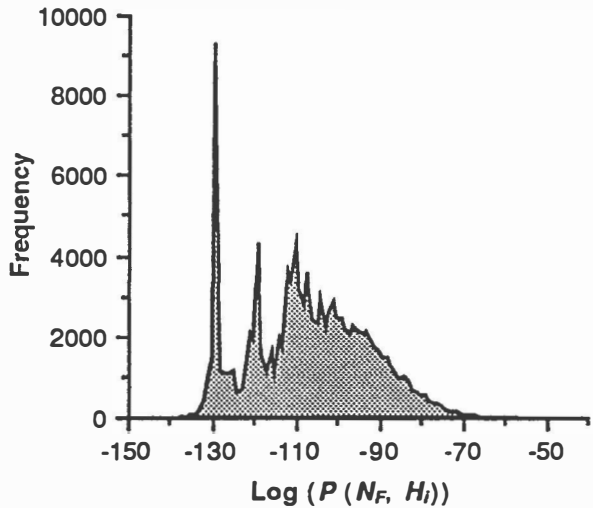

Figure 9 A distribution of the log of the joint probabilities $P(N_F, H_i)$ of the hypotheses $H_i$ instantiated during the S/NSI simulation on the findings $N_F$ of CPC1.

We also recorded the time during simulation at which each hypothesis was instantiated. In Figure 11 appears the distribution as a function of time of the log of $P(N_F, H_i)$ for the hypotheses $H_i$ generated by REF, where $N_F$ are from CPC1. Early in the REF simulation, the variance of the $P(N_F, H_i)$ values was large, and the mean of the distribution was relatively small. However, soon after self-importance sampling begins at trial 20,000 (after approximately 1/20 of the total simulation time had elapsed), the variance of the distribution decreased, and the mean increased substantially. By contrast, the distribution (as a function of time) of the $P(N_F, H_i)$ values produced by S/NSI on CPC1 retained the general shape of the distribution in Figure 9 over the entire period of simulation, since the importance distribution of S/NTTB did not change as the simulation progressed.

Although we limit our discussion of the distributions of $P(N_F, H_i)$ to CPC1, we observed similar results on the corresponding distributions from CPC2.

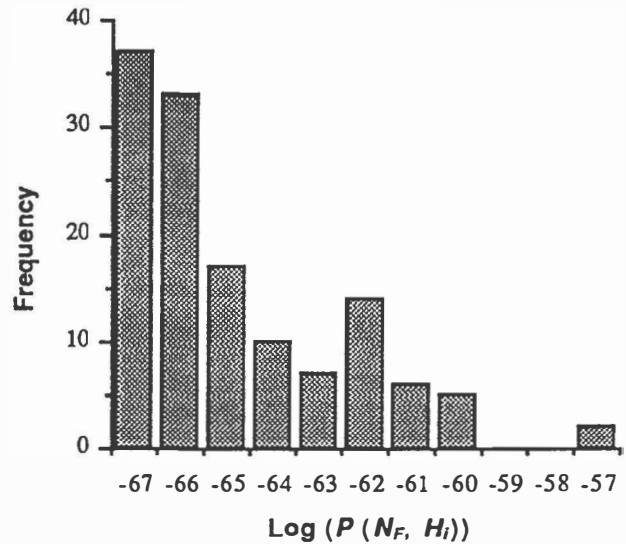

Figure 10 An enlargement of the right-hand side of the distribution that appears in Figure 8.

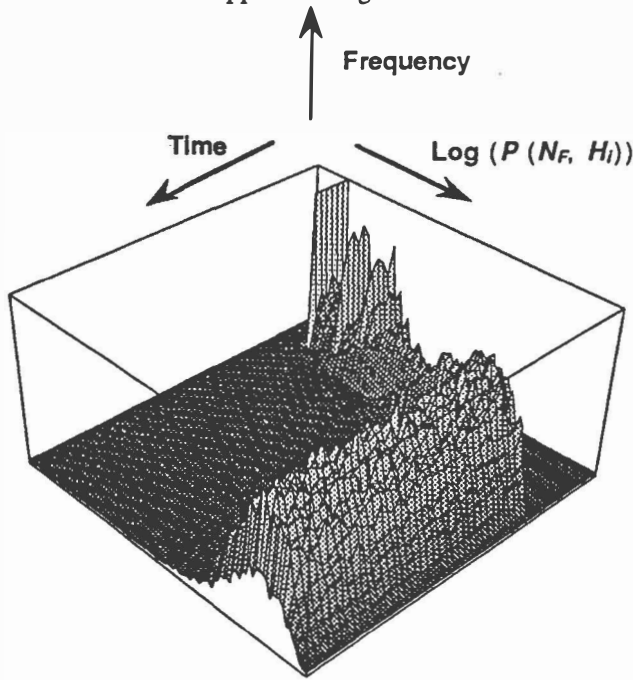

Figure 11 A plot as a function of time of the distribution of Log $(P(N_F, H_i))$ of the hypotheses $H_i$ instantiated during the REF simulation on CPC1.

We recorded the average number of diseases present in a disease hypothesis $H_i$ as a function of the log of the joint probability $P(N_F, H_i)$. Figures 12 and 13 display this distribution for the REF runs on CPC1 and CPC2, respectively. Note that the distributions in Figures 12 and 13 represent the hypotheses instantiated by the simulation, not the actual distribution of disease hypotheses.

Observe that, in both Figures 12 and 13, the instantiations of the QMR-DT belief network with the largest joint probabilities contained approximately 10 diseases. This value is quite large, given that CPC1 had five diseases in its pathological diagnosis and CPC2 had three. We are currently investigating the reasons that the QMR-DT model exhibits this behavior. In particular, the noisy-OR gate may be responsible for this behavior.

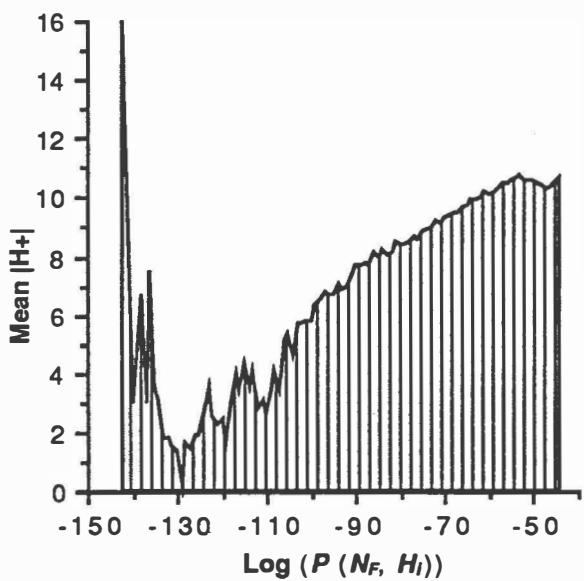

Figure 12 A plot of the mean cardinality of $H^+$ (the number of diseases present in disease hypothesis $H$) as a function of the log of the joint probability $P(N_F, H_i)$ for hypotheses instantiated by REF on CPC1.

In this section, we discussed and depicted various distributions as a function of the joint probabilities of the QMR-DT belief network during simulation. For the sake of brevity, we do not show the corresponding distributions as a function of the sample score. We found a corresponding similarity in the distributions of sample scores generated by REF and S. Also, the distribution of the sample scores generated by S/NSI contained a number of outliers of high probability. On both CPC1 and CPC2, we observed that the largest sample scores were generated by hypotheses with approximately 10 diseases.

In summary, we observed that successive runs of likelihood-weighting simulation with importance sampling, self-importance sampling, and Markov blanket scoring were able to produce similar estimates of posterior marginal probabilities of disease, when the simulation was presented with two difficult diagnostic cases. This reproducibility supports our belief that the estimates of the reference distributions have converged appreciably to the posterior probabilities implied by the QMR-DT model. The heuristic ITB algorithm provides the simulation with a set of likely diseases given the findings observed; however, the convergence of the simulation algorithm does not depend on the heuristic. Rather, the simulation appears to be more sensitive to the self-importance sampling and Markov blanket scoring.

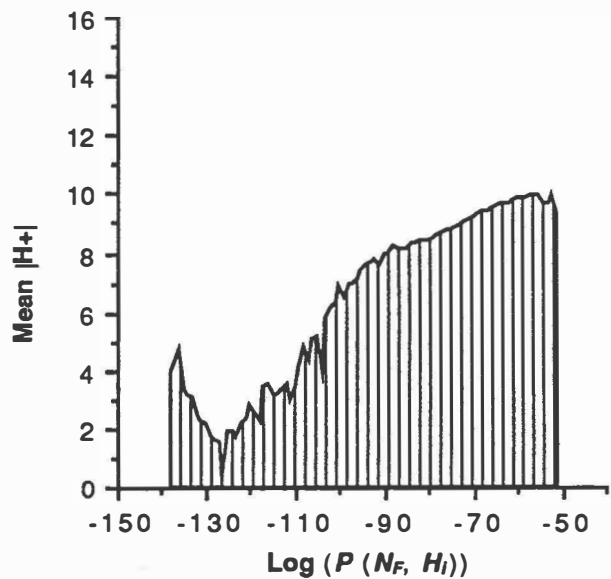

Figure 13 A plot of the mean size of $H^+$ (the number of diseases present in disease hypothesis $H$) as a function of the log of the joint probability $P(N_F, H_i)$ for hypotheses instantiated by REF on CPC2.

The simulation also provides us with insight on the behavior of the underlying model. For example, we found that the instantiations of the network with the largest joint probability had approximately 10 diseases. We plan to investigate further the behavior of the model using the results of simulation.

## 6. Discussion

On two large diagnostic cases with multiple diseases, we observed, after 960 minutes of simulation, that the posterior estimates of the S algorithm correlated well with the estimates of the reference distribution. In a previous study of the behavior of S/NMBS on smaller test cases with single-disease diagnoses, we observed appreciable convergence within 2 hours of simulation time on a Macintosh IIci [Shwe, Middleton, et al., 1990]. Since the simulation is readily amenable to parallelization, we do not believe that the current running time of our serial implementation of the S algorithm on a personal computer will be a long-term limitation for practical applications.

Both the Markov blanket scoring component and the self-importance sampling heuristic of the S algorithm are particularly suited to the two-level connectivity of the current QMR-DT belief network. Because we have, for the time being, assumed diseases to be marginally independent, the Markov blankets of the disease nodes are relatively small, to the extent that the time needed to compute the Markov blanket probability of all the disease nodes in the network is comparable to the time required to compute the sample score of an instantiation of the network.

Because we assumed marginal independence, we could perform self-importance updating by simply sampling the diseases based on the the simulation's current estimates of their posterior marginal probabilities. If we were to introduce disease-to-disease dependencies to the QMR-DT model, the self-importance sampling would require that we sample a disease node based on the state of that node's parents in the current instantiation of the network. In general, this requirement necessitates additional storage and computation to provide estimates of the conditional probability of a node.

The convergence results that we present in this paper demonstrate that likelihood-weighting simulation is a viable method of inference for the two-level QMR-DT belief network. Moreover, the results provide promise that likelihood-weighting simulation will be a useful inference tool not only on future versions of the QMR-DT belief network, which will have richer sets dependencies, but also on other large multiply connected belief networks, for which exact inference algorithms are not practical.

## Acknowledgments

We are grateful to other members of the QMR-DT project group—David Heckerman, Max Henrion, Eric Horvitz, Harold Lehmann, and Blackford Middleton—who were instrumental in developing the QMR-DT model. We thank Ross Shachter for providing insight on the likelihood-weighting simulation and for encouraging us to implement self-importance sampling and Markov blanket scoring. Randy Miller graciously provided us with the INTERNIST-1 knowledge base. Lyn Dupré provided valuable comments on drafts of this paper.

This work was supported by the National Science Foundation under Grant IRI-8703710 and the U. S. Army Research Office under Grant P-25514-EL. Computing facilities were provided by the SUMEX-AIM Resource under the National Library of Medicine Grant LM05208.

## References

1. Castleman B, Scully RE, McNeely BU. Case records of the Massachusetts General Hospital: Weekly clinicopathological exercises: Case 21-1972. New England Journal of Medicine 1972; 286: 1146-1153.

2. Cooper GF. The computational complexity of probabilistic inference using Bayesian belief networks. Artificial Intelligence 1990; 42: 393–405.

3. Cryer PE, Kissane JM. Clinicopathologic conference: pulmonary hypertension. American Journal of Medicine 1974; 69: 127-134.

4. Fung R, Chang KC. Weighting and integrating evidence for stochastic simulation in Bayesian networks. In: Henrion M, ed. *Proceedings of the Fifth Workshop on Uncertainty in Artificial Intelligence*. Windsor, Ontario: 1989: 112–117.

5. Heckerman DE. A tractable inference algorithm for diagnosing multiple diseases. In: Henrion M, ed. *Proceedings of the Fifth Workshop on Uncertainty in Artificial Intelligence*. Windsor, Ontario: 1989: 174-181.

6. Henrion M. Towards efficient probabilistic diagnosis in multiply connected networks. In: *Proceedings of the Conference on Influence Diagrams for Decision Analysis, Inference, and Prediction*. Berkeley, CA: 1988.

7. Lawrence L. Detailed diagnosis and surgical procedures for patients discharged from short-stay hospitals. Vital & Health Statistics 1986; Series 13: No. 82.

8. Miller RA, Pople HEJ, Myers JD. Internist-1: An experimental computer-based diagnostic consultant for general internal medicine. New England Journal of Medicine 1982; 307: 468-476.

9. Pearl J. Evidential reasoning using stochastic simulation of causal models. Artificial Intelligence 1987; 32: 245-257.

10. Pearl J. *Probabilistic Reasoning in Intelligent Systems: Networks of Plausible Inference*, San Mateo, CA: Morgan Kaufman, 1988.

11. Rubinstein RY. *Simulation and the Monte Carlo Method*, John Wiley & Sons, 1981.

12. Shachter RD, Peot M. Simulation approaches to general probabilistic inference on belief networks. In: Henrion M, ed. *Proceedings of the Fifth Workshop on Uncertainty in Artificial Intelligence*. Windsor, Ontario: 1989: 311-318.

13. Shwe MA, Middleton BF, Heckerman DE, Henrion M, Horvitz EJ, Lehmann HP, Cooper GF. *Probabilistic Diagnosis Using a Reformulation of the INTERNIST-1/QMR Knowledge Base. Knowledge Systems Laboratory Memo no. KSL-90-09*. Stanford CA: Stanford University, 1990.